\documentclass[conference]{IEEEtran}
\IEEEoverridecommandlockouts
\usepackage{cite}
\usepackage{amsmath,amssymb,amsfonts}
\usepackage{algorithmic}
\usepackage{graphicx}
\usepackage{textcomp}
\usepackage{subcaption}
\usepackage{subfloat}
\usepackage{threeparttable}
\usepackage{xcolor}
\usepackage{multirow}
\usepackage{booktabs}
\usepackage{makecell}
\usepackage{amsthm}
\usepackage{algorithmic}
\usepackage{hhline}
\usepackage[ruled,linesnumbered]{algorithm2e}
\def\BibTeX{{\rm B\kern-.05em{\sc i\kern-.025em b}\kern-.08em
    T\kern-.1667em\lower.7ex\hbox{E}\kern-.125emX}}
\begin{document}

\title{DanZero: Mastering GuanDan Game with Reinforcement Learning
}
\author{Yudong Lu,
		Jian Zhao,
		Youpeng Zhao,
		Wengang Zhou,
		Houqiang Li
}
\maketitle

\begin{abstract}
Card game AI has always been a hot topic in the research of artificial intelligence.
In recent years, complex card games such as Mahjong, DouDizhu and Texas Hold'em have been solved and the corresponding AI programs have reached the level of human experts.
In this paper, we are devoted to developing an AI program for a more complex card game, GuanDan, whose rules are similar to DouDizhu but much more complicated.
To be specific, the characteristics of large state and action space, long length of one episode and the unsure number of players in the GuanDan pose great challenges for the development of the AI program. 
To address these issues, we propose the first AI program DanZero for GuanDan using reinforcement learning technique.
Specifically, we utilize a distributed framework to train our AI system.
In the actor processes, we carefully design the state features and agents generate samples by self-play. 
In the learner process, the model is updated by Deep Monte-Carlo Method. 
After training for 30 days using 160 CPUs and 1 GPU, we get our DanZero bot.
We compare it with 8 baseline AI programs which are based on heuristic rules and the results reveal the outstanding performance of DanZero.
We also test DanZero with human players and demonstrate its human-level performance.

\end{abstract}

\section{Introduction}
As a good benchmark for measuring the strength of artificial intelligence (AI), games have been attracting lots of research efforts in the field of machine learning, especially reinforcement learning.
Thanks to the rapid development of reinforcement learning, recent years have witnessed significant progress of AI in various games, including board games such as Go and chess \cite{silver2016mastering, silver2017mastering, silver2018general}, card games such as Texas Hold'em and Mahjong\cite{ heinrich2016deep, brown2018superhuman, moravvcik2017deepstack, li2020suphx} and video games such as StarCraft and DOTA \cite{vinyals2019grandmaster, berner2019dota}.
However, games with imperfect information and large state and action space still pose a challenging issue for reinforcement learning.

\begin{figure}[t]
	\centering
	\subfloat[The information set size and count of different games]{
		\includegraphics[width=0.4\textwidth]{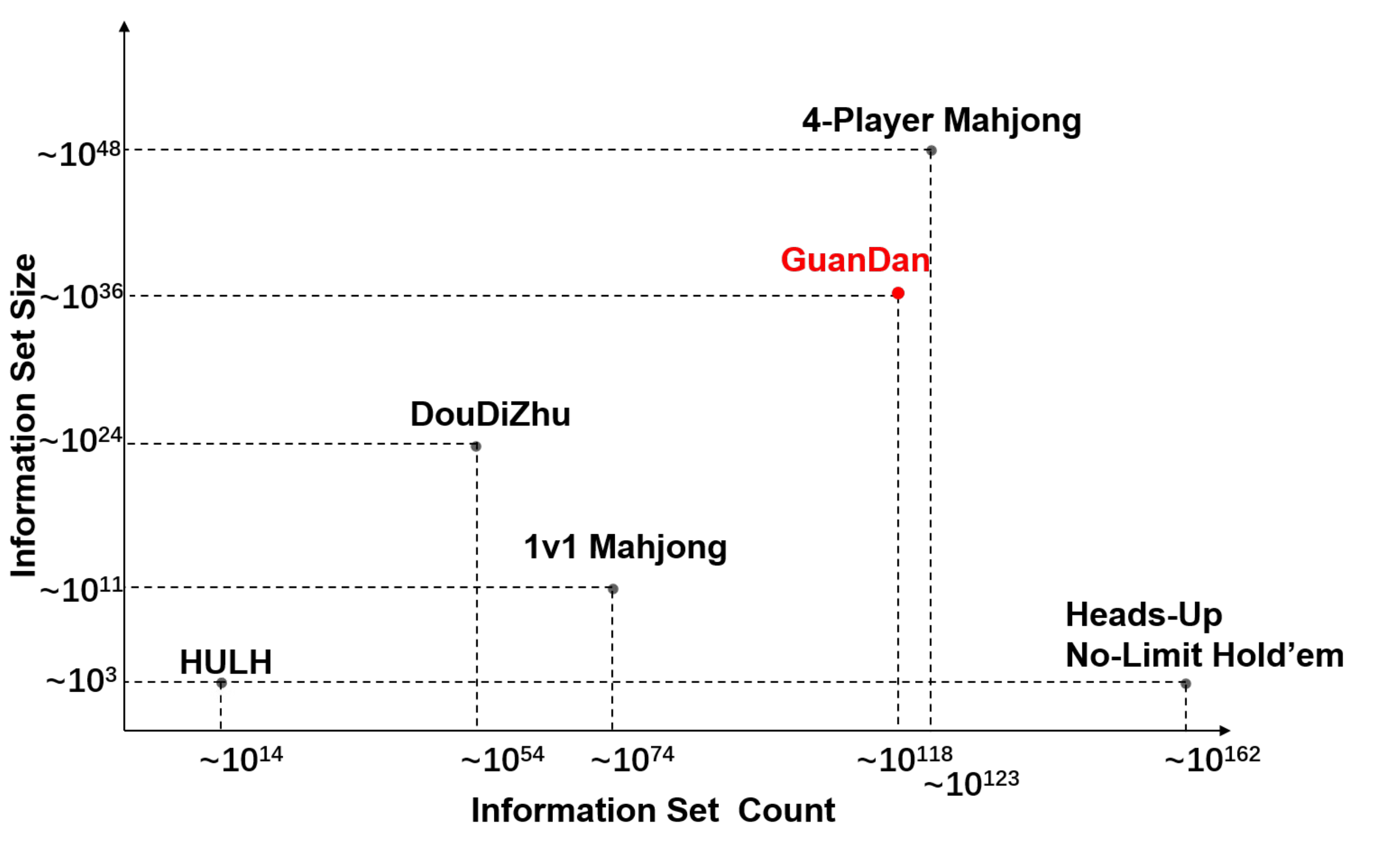}
		\label{fig1-1}
	}
	\\
	\subfloat[The size of action space in different games]{
		\includegraphics[width=0.4\textwidth]{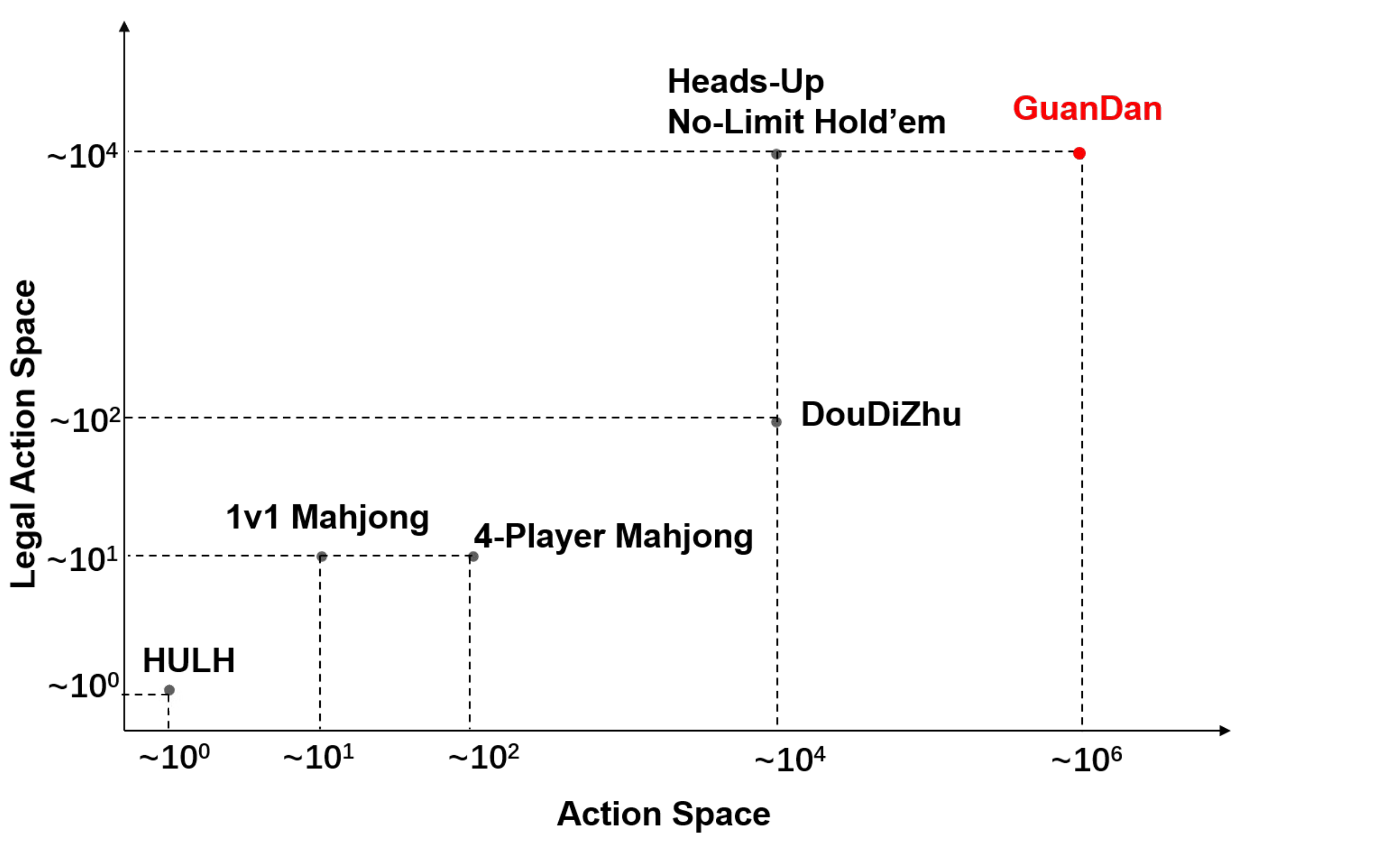}
		\label{fig1-2}
	}
	\caption{Game complexity of some imperfect-information games, including Heads-Up Limit Texas Hold'em (HULH), Heads-Up No-Limit Texas Hold'em, 1v1 Mahjong, 4-Player Mahjong, DouDizhu and GuanDan. }
	\label{fig1}
\end{figure}

In this work, we are dedicated to developing an AI program for another card game, which is called GuanDan and still under-explored. This game is similar to another complex imperfect-information game, DouDizhu, in some characteristics in that both of them involve cooperation and competition simultaneously under a partially observable environment and possess large state and action space.
However, GuanDan is much more complicated than DouDizhu and has larger state and action space, which is shown in Figure~\ref{fig1}.
It can be observed that among several common imperfect-information games, the information set size and count of GuanDan are comparatively equivalent to those of 4-player Mahjong and much higher than other games in comprehensive consideration.
Whereas, the size of action and legal action space of GuanDan is up to $10^{6}$ and $10^{4}$, respectively, much larger than that of other games listed in the Figure. Although DouDizhu has been studied a lot recently and some progress has been achieved \cite{you2020combinatorial, jiang2019deltadou, zha2021douzero, zhao2022douzero+},  developing an AI program for GuanDan is very challenging and remains unsolved.

Compared to DouDizhu, GuanDan is much more challenging in several aspects:
\begin{itemize}
    \item \textbf{The state and action space of GuanDan is much larger:} In a GuanDan game, there are two decks of pokers used while DouDizhu only uses one deck of cards. 
    What's more, the suits of cards are important in GuanDan as the combinations of cards are affected by this factor while suits of cards are usually ignored in DouDizhu.
    In addition, there exist ``wild card'' and ``level card'' in GuanDan, making it much more difficult to play compared with DouDizhu.
    \item \textbf{The length of one episode of GuanDan game is long:} GuanDan involves a concept of ``leveling up'', and the game ends only when a team of players have upgraded over level A so that one episode of game contains several rounds.
    In fact, each agent in GuanDan has to make over 100 decisions in each episode while one agent in DouDizhu only needs to make about 10 decisions in one episode.
    \item \textbf{The number of players in GuanDan game may change as the game progresses:} There are four players in GuanDan, two of whom make up a team against the team of the remaining two players.
    One round of GuanDan game terminates only when both players of any team have emptied their hand cards, making this game very complex.
    For example, if one player has emptied his hand cards, the following game will be turned into an unequal situation, \emph{i.e} two players cooperate against the rest one.
    In contrast, a DouDizhu game ends as long as any player empties his hand cards so that this game pattern is fixed.
\end{itemize}

Almost all of the existing AI programs for GuanDan are based on heuristic rules and many carefully designed techniques are applied to deal with different situations.
For example, a player may give priority to playing cards that are small or can not form special card types when leading the first trick.
When a player plays cards passively to cover cards of other players, using cards of the same card type is prior to playing bombs. 
There's one work \cite{shen2020imperfect} that tried to develop an AI program for GuanDan which adopts Upper Confidence Bound Apply to Tree (UCT) algorithm but it performs only slightly better than random agents, \emph{i.e.}, agents choose actions randomly from legal action set .

In this work, we propose an AI system for GuanDan called DanZero using reinforcement learning techniques.
Considering the large state and action space, classical value-based reinforcement learning algorithms such as Deep Q-Learning (DQN) \cite{mnih2015human} will probably suffer from overestimating issue \cite{zahavy2018learn}. 
Similarly, policy gradient methods such as A3C \cite{mnih2016asynchronous} also perform unsatisfactorily in issues with large action space as they fail to leverage action features.
In this way, we choose to adopt Monte Carlo method enhanced by neural networks, which can utilize the action features and approximate true values without bias \cite{sutton2018reinforcement}.
What's more, we adopt feature encoding techniques to process the state and action features and implement a distributed self-play reinforcement learning framework to facilitate training. 
Integrating these techniques, our AI system for GuanDan is able to beat all other existing algorithms, proving the effectiveness of our methods.

\section{Related Work}
In this section, we describe some background on imperfect-information games and the application of reinforcement learning for developing the corresponding game AI.

Compared to perfect-information games, the agent of imperfect-information games need to handle hidden information and randomness, which is more close to real world as there always exist stochasticity and unknowns.
In this way, imperfect-information games always provide more challenging and important research questions. 
In the classical imperfect-information games, poker games, Counterfactual Regret Minimization (CFR) \cite{neller2013introduction} and its variants where a model of game is required to traverse the game tree during computation are often adopted.
To handle a large-scale imperfect-information game, learning an abstraction of state or action space to reduce the game to a manageable size is often applied \cite{bowling2015heads, moravvcik2017deepstack, brown2019deep}.
However, as GuanDan involves both cooperation and competition at the same time and the number of players may change as the game goes on, such a complex setting poses great challenges to these classical algorithms in poker games.
Although utilizing deep neural networks to generalize across states helps CFR methods obviate the need for abstractions \cite{li2018double, steinberger2019single}, this family of algorithms still have difficulty dealing with games with large state and action space.

Different from CFR methods which rely on game-tree traversals, reinforcement learning can help models learn skills through interactions with the environment so that this technique is very suitable for large-scale games. In fact, thanks to the development of reinforcement learning, there is recently a growing trend in utilizing this technique to solve imperfect-information games. For instance, AI programs for famous large-scale games such as DOTA, StarCraft and Honor of King have been developed and achieved amazing performance \cite{berner2019dota, vinyals2019grandmaster, ye2020mastering}. As for card games, reinforcement learning has also been successfully adopted in Mahjong, Texas Hold'em, DouDizhu and so on \cite{li2020suphx, heinrich2016deep, brown2018superhuman, you2020combinatorial, zha2021douzero}. What's more, reinforcement learning can be combined with many other techniques such as search \cite{brown2020combining} and opponent modeling \cite{he2016opponent, knegt2018opponent} and shows satisfactory performance. Considering the advantages of this powerful technique, we choose reinforcement learning to develop an AI program for the unsolved GuanDan game.

\section{Basic Rules of GuanDan}
\begin{figure}[t]
	\centering
	\includegraphics[width=1\columnwidth]{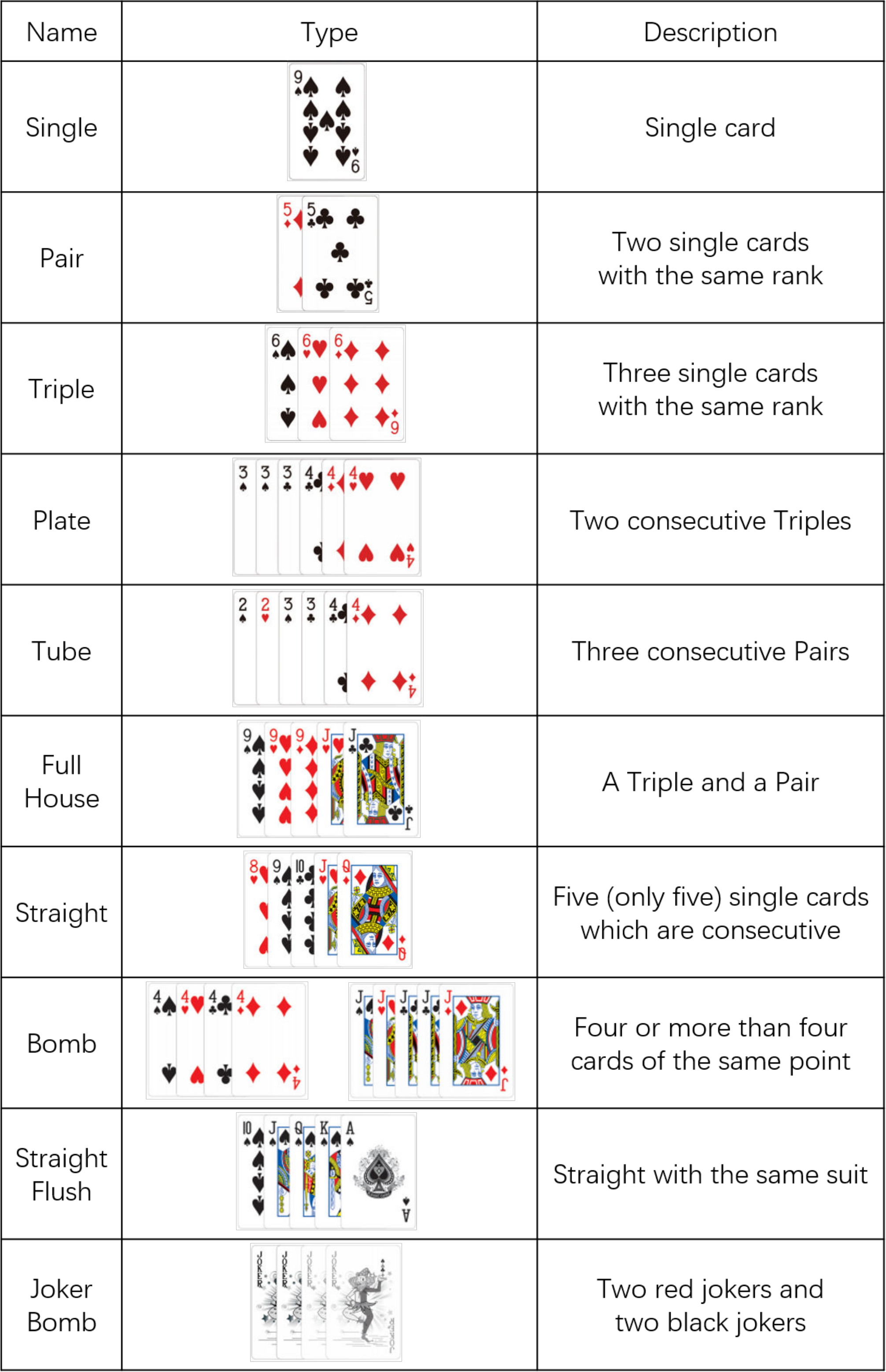}
	\caption{A list of all card types in GuanDan.}
	\label{fig2}
\end{figure}

In this section, we give a brief introduction to the basic rules of GuanDan. The card types of this game are listed in Figure~\ref{fig2} .

There are four suits in the cards used in GuanDan game, including Hearts (H), Spades (S), Diamonds (D) and Clubs (C). The basic rank of single cards is, from high to low, Red Joker (RJ), Black Joker (BJ), Level cards, A, K, Q, J, 10, 9, 8, 7, 6, 5, 4, 3, 2. When forming Tubes, Plates, Straights or Flushes, Aces can be seen as 1, just below 2. For Solos, Pairs, Triples, Tubes, Plates, Full Houses and Straights, cards can only cover the same type. For Full Houses, they are ranked by the points of Triple part. Bombs can cover these mentioned combinations and Bombs with more cards can cover Bombs with fewer cards. If the numbers of cards are the same between two Bombs, they are also ranked by their points. Flush Straights can cover Bombs with less than six cards and the relationship between this kind of card type is determined by the points. Finally, Joker Bombs can beat any card types.

In a GuanDan game, there are two decks of standard pokers used, including Jokers, and four players sitting around a square table, each of whom has 27 cards in hand. 
The players sitting opposite each other belong to the same camp.  What's more, there exists the concept of ``leveling up'', ``level cards'' and ``wild cards''.
To be specific, both camps of this game own their own level which starts from 2 to A. 
The first round of a GuanDan game is always played at Level 2 and the levels of the subsequent rounds are determined by the level of the camp who have won in the previous round.
Cards of the same rank as the level of current round are called ``level cards'' and they rank just below Jokers when being played singly.
This kind of cards can also be used at their ranks in natural order when making up other combinations.
In addition, level cards in Heart are ``wild cards" and they can be utilized in place of any cards needed to make up a combination except for Jokers. 
The camp which first levels up over level `A' will win the game. To this end, one GuanDan game usually contains several rounds.

Players play cards in counterclockwise order and the player leading the first trick can play any type of cards from his hand.
The other players can play cards of the same type or bombs, which are larger than cards played by the previous player, or they can choose to pass.
A trick continues until three players pass in succession and the player who played the last cards leads in the next trick.
Such procedures repeat until three players have no card left or players of the same camp have emptied their cards and this round ends.
The first one emptying the cards is called the Banker, and other players are called the Follower, the Third and the Dweller according to the order of their emptying the cards, respectively.
Only the team of the Banker can promote the level and the promoted number can be three or two or one according to the partner of the Banker.
If the winning team manages to promote three levels, their opponents are called the Double-Dweller.
In addition, if one trick ends after the Banker or the Follower finishes the cards, his partner will lead the next trick. From the second round on, before the first trick begins, the Dweller of the previous round has to pay Tribute to the Banker by giving his biggest single card other than the wild card.
In return, the Banker needs to give a single card back with a point not higher than 10.
Then the Dweller will lead the first trick. If there is Double-Dweller in the previous, both of this camp have to pay the Tribute and the Banker accepts the higher ranked Tribute.
The winners of the last round also need to return cards as discussed above.
However, when the player or the team that need to pay Tribute have two Red Jokers in hand, the Tribute phase can be canceled and the Banker will lead the first trick.

Last but not least, when the level of current round is Q or K and the winner can promote 3 or 2 levels, the level of A can not be skipped. At the round with level A, a camp can only win when the Banker's partner is the Follower or the Third. This section just introduces brief rules of GuanDan and more detailed rules can be found in Wiki. \footnote{https://en.wikipedia.org/wiki/Guandan\#Playing}

\section{Method}
In this section, our DanZero will be described in two aspects: the model architecture design and the training algorithm.

\subsection{Model Architecture Design}
\begin{figure}[t]
	\centering
	\includegraphics[width=1\columnwidth]{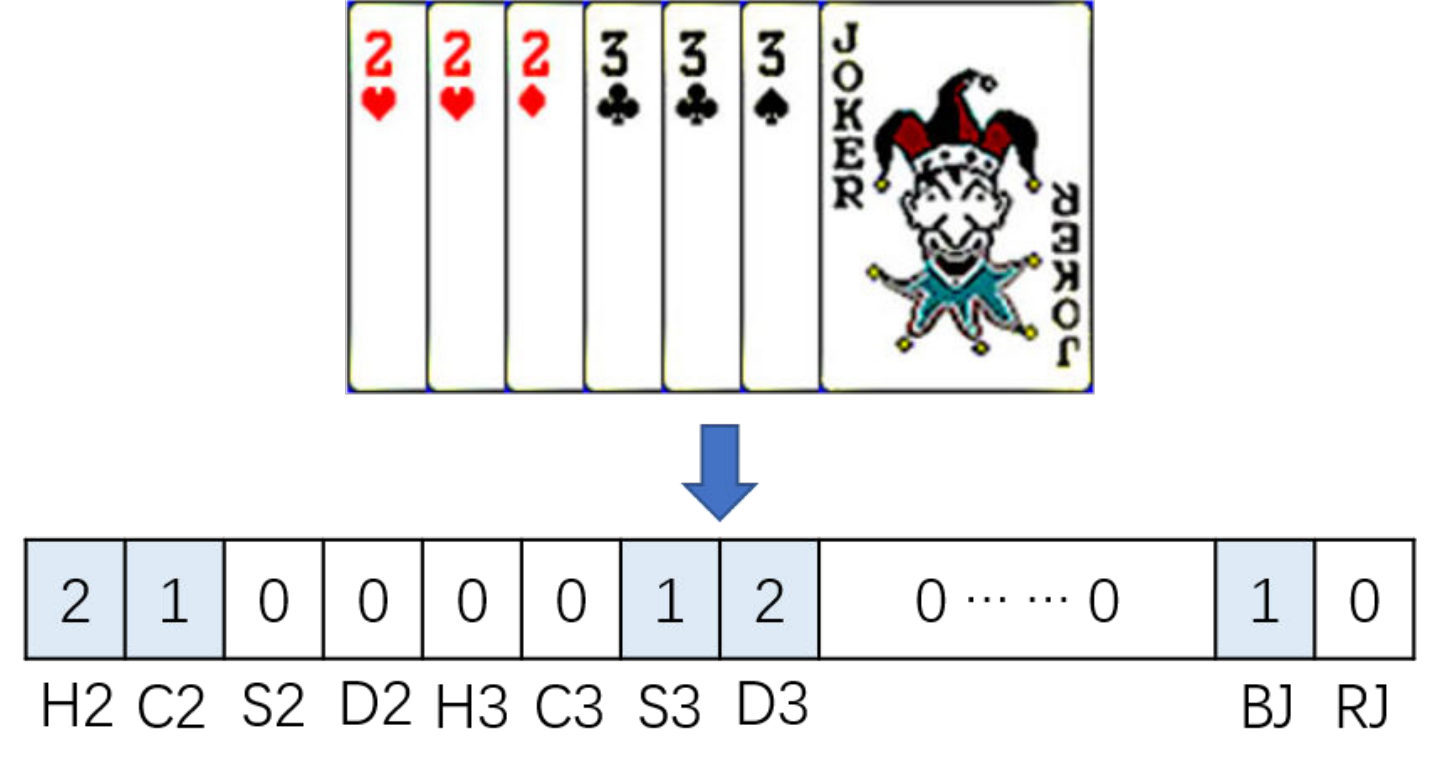}
	\caption{An example to show the encoding of one hand. `H' is the abbreviation of Heart. `C' is the abbreviation of Club. `S' is the abbreviation of Spade and `D' is the abbreviation of Diamond. `BJ' is the abbreviation of Black Joker and `RJ' is the abbreviation of Red Joker.
	0, 1, 2 represents the number of such cards.}
	\label{fig3}
\end{figure}

The key in our model architecture design is taking all relevant information and candidate action as input and outputting the state-action value.
We encode each card combination with a 54-dimensional vector, corresponding to the 54 cards in poker.
There are 3 possible values for each element in the vector, \emph{i.e.} $\{ 0, 1, 2 \}$, indicating the number of cards of the corresponding suits and points.
An example is shown in Figure~\ref{fig3}.
In this way, the feature encodes the suit information, and the dimension size is acceptable

The feature of the state is composed of a vector with 513 dimensions, and their physical meanings are listed as follows ( from the view of one player ):
\begin{itemize}
    \item $\left[ 0 - 53 \right]$: our current hand. 
    \item $\left[ 54 - 107 \right]$: the remaining cards, \emph{i.e.}all the cards except our current hand and all played cards.
    \item $\left[ 108 - 161 \right]$: the last move and the cards that we are going to play must be able to cover this combinations of cards. If we have to lead the trick, these dimensions are set to be zero. 
    \item $\left[ 162 - 215 \right]$: the last move of partner. If the last move of a teammate is ``pass'', this vector is set to be zero. If the partner has finished his hand cards, these dimensions are set to be -1.
    \item $\left[ 216 - 299 \right]$: the number of remaining cards of other three players which is recorded in the order of playing cards.
    \item $\left[ 300 - 461 \right]$: the played cards of other three players which is recorded in the order of playing cards.
    \item $\left[ 462 - 501 \right]$: the level of our team and opponent team. 
    \item $\left[ 501 - 513  \right]$: flag for wild cards, namely, whether we have wild cards in hand and whether these cards can make up Bombs, Straight Flushes, Straights, or other card types except Single and Joker Bomb.
\end{itemize}

\begin{figure*}[t]
	\centering
	\includegraphics[width=0.94\textwidth]{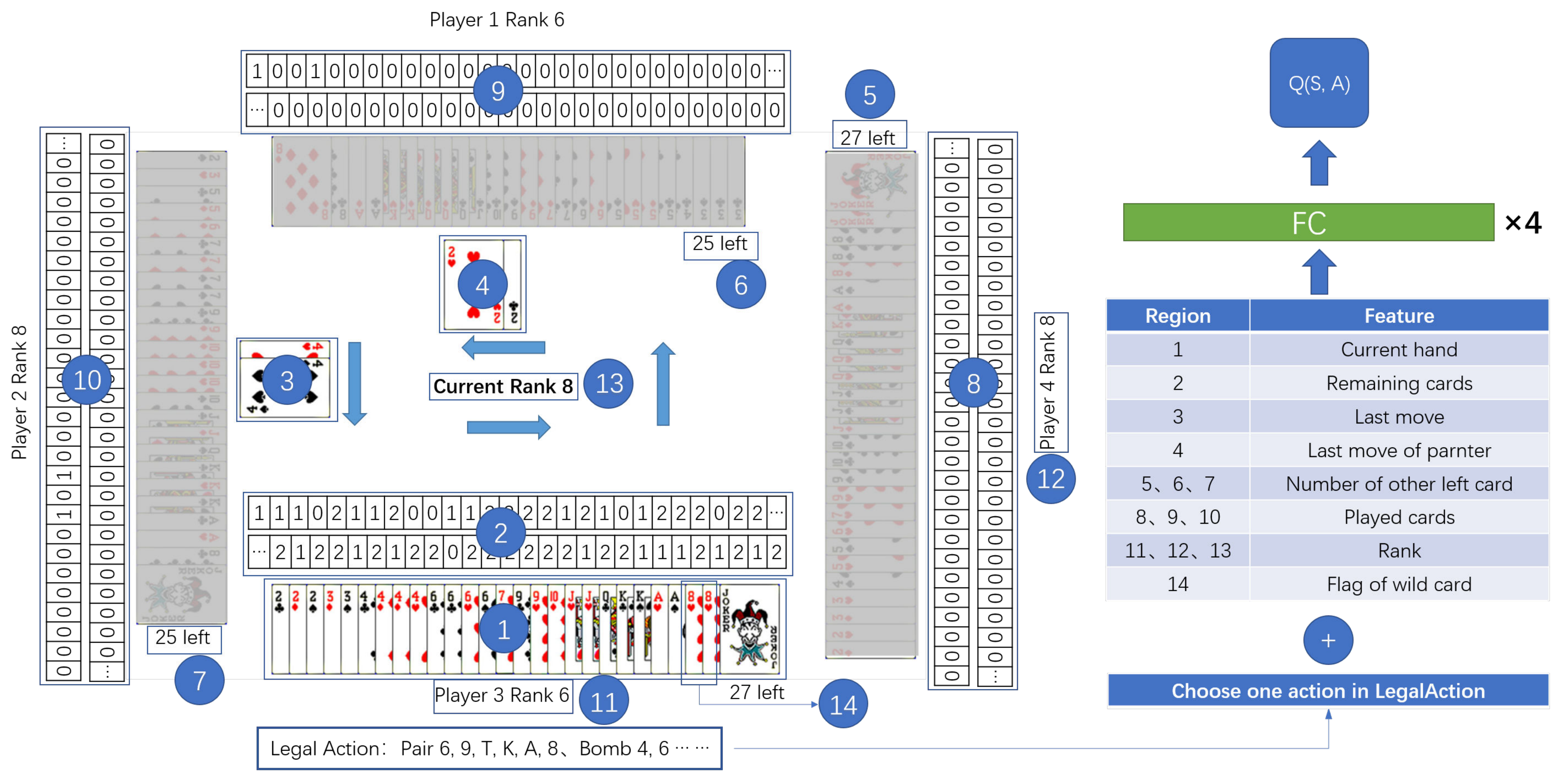}
	\caption{An example to show different regions in the GuanDan game for better understanding of state features. The right part indicates the architecture of the network, which takes state features and legal actions as input and outputs the Q-value of a state-action pair.  }
	\label{fig4}
\end{figure*}

As for the action features, they are also represented with a 54-dimensional vector.
The network that we adopt consists of several layers of Multi-Layer Perception (MLP) and the input is the concatenation of the state and action features, which is a 567-dimensional vector.
The output of the network is the Q-value for one state-action pair. Figure~\ref{fig4} demonstrates how we divide different regions to form the state vector and the architectures of the network that we adopt.

\subsection{Distributed Q-learning}

In this subsection, we introduce the training process of our method. 
Our DanZero is trained with deep Monte Carlo method and self-play procedure.
The overall framework consists of two parts: actor and learner.

\subsubsection{Actor}

The actor is responsible for the game simulation and information collection. In each iteration, each actor receives the latest model parameter from the learner.
Then, the environment initializes a new episode.
At each time step, each agent $i$ receives the last moves of other three players $a_{-i}$ and updates the feature $\tau_i$.
According to the current state, we first calculate the legal actions set $A$.
The model takes $\tau$ and each legal action $a$ as input and outputs the state-action value $Q(\tau_i,a)$.
Then action is selected by $\epsilon$-greedy policy.
Specifically, an action is randomly selected from the legal action set with the probability of $\epsilon$, and there is also a probability of $1-\epsilon$ to choose the action with the largest $Q(\tau_i,a)$. 
Each agent takes turns according to the above operation until an episode ends.

After the end of one round in a GuanDan game, we assign a value for every sample according to the result of this round.
To be specific, for the winning team, samples of their trajectory will be assigned +3, +2 and +1 when the partner of the Banker is the Follower, the Third and the Dweller, respectively.
And the samples of losing camp will be assigned corresponding negative value. When the level of the last round of one game is A, the reward changes to 0 when the Banker's partner is not the Dweller as they can not win the game in this case.
The agent trajectory data tuple $(\tau, a, Q(\tau,a), r)$ is sent to the learner to train the model after one episode terminates. Considering the self-play procedure, each episode will produce 4 trajectories.

Need to add that, in the second and subsequent round of one episode, there exists a Tribute phase. Because the logic of Tribute is quite different from playing cards, we utilize heuristic rules to make decisions in this phase. What's more, data in this phase will not be saved. Detailed rules are available in supplement materials.

\subsubsection{Learner}

The learner is responsible for network update.
In each iteration, the learner receives the collected episode data from the actors and the data is stored in a buffer.
The learner samples a batch of data from the buffer to update the network with Deep Monte Carlo.
Deep Monte Carlo is an effective value-based algorithm especially in such episodic and reward-sparse tasks which has a very large state and action space.
However, considering the transportation delay in distributed reinforcement learning, it is necessary to preprocess the state-action value sent by the actor as the following equation shows.
\begin{equation}\label{Target}
Q_p(\tau,a) = clip(\frac{Q(\tau,a;\theta_{l})}{Q(\tau,a;\theta_{a})}, 1 - \lambda, 1 + \lambda) * Q(\tau,a;\theta_{a}),
\end{equation}
where $Q(\tau,a;\theta_{a})$ refers to the state-action value predicted by the $\theta$ parameter on the actor side, and $Q(\tau,a;\theta_{l})$ refers to the state-action value predicted on the learner side.
$\lambda$ is a hyperparameter and the definition of clip function is shown as follows:
\begin{equation}\label{eqfun}
clip(x, x_{min}, x_{max}) = \left\{
\begin{array}{ll}
x_{max},  & x > x_{max}\\
x,      & x_{min}\leq x\leq x_{max}\\
x_{min},   & x < x_{min} \\
\end{array}
\right. .
\end{equation}
Using the above preprocessing formula can remove the samples with a large distribution difference against the model on the learner side, so that the given target value is time-sensitive.
Then we construct an optimization function as follows:
\begin{equation}\label{loss}
Loss= \frac{1}{N} \sum_{i=i}^N [ Q_p(\tau,a)  - r)]^2, 
\end{equation}
where $N$ represents the batch size.

By adopting the Distributed Q-Learning, we can parallelize multiple actor processes so that the training process of our AI system is efficient. The overall algorithm framework is summarized in Algorithm~\ref{actor} and \ref{learner}.

\begin{table*}[htbp]
    \centering
    \resizebox{\linewidth}{!}
    {
        \renewcommand{\arraystretch}{1.25}
        \begin{tabular}{|c|c|c|c|c|c|c|c|c|c|c|}
        \hline
            Win Rate ($\%$) & baseline8 & baseline7 & baseline6 & baseline5 & baseline4 & baseline3 & baseline2 & baseline1 & Ours- & Ours \\ \hline
baseline8 & -   & 26.72    & 1.14         & 100.00         & 0.00        & 13.10         & 0.00       & 0.00         & 0.00        & 0.00         \\ \hline
baseline7 & 73.28   & -    & 42.52         & 100.00         & 6.93        & 15.31         & 0.00       & 0.00        & 0.00         & 0.00        \\ \hline
baseline6 & 98.86   & 57.48    & -        & 97.67        & 18.40        & 47.96        & 28.19        & 15.64         & 0.00        & 0.00        \\ \hline
baseline5 & 0.00   & 0.00    & 2.33         & -         & 0.00         & 0.00         & 0.00         & 0.00         & 0.00         & 0.00         \\ \hline
baseline4 & 100.00  & 93.07   & 81.60        & 100.00        & -        & 83.15         & 36.42       & 55.71        & 17.32        & 12.55       \\ \hline
baseline3 & 86.90   & 84.69    & 52.04        & 100.00        & 16.85         & -        & 12.89       &  14.28       & 0.00        & 0.00        \\ \hline
baseline2 & 100.00  & 100.00   & 71.81        & 100.00         & 63.58        & 88.19        & -       & 54.44        & 25.33       & 17.39       \\ \hline
baseline1 & 100.00  & 100.00   & 86.36         & 100.00        & 54.29        & 86.78        & 45.56       & -        & 12.77       & 9.82       \\ \hline
Ours-      & 100.00  & 100.00    & 100.00        & 100.00        & 82.68       & 100.00        & 74.67       & 87.23       & -       & 46.55       \\ \hline
Ours       & \textbf{100.00}   & \textbf{100.00}   & \textbf{100.00}       & \textbf{100.00}        & \textbf{87.45}       & \textbf{100.00}        & \textbf{82.61}       & \textbf{90.12}       & \textbf{53.45}       &  -       \\ \hline
        \end{tabular}
    }
    \caption{The average performance of the compared algorithms by playing 1000 episodes of GuanDan. The win rate of each row is achieved by test between the bot in the first column against other algorithms. ``Our-'' represents the abrasive model that removes the flag for wild cards. The results of ``Our'' and ``Our-'' are achieved after training for 30 days.}
    \label{tab1}
\end{table*}

\begin{algorithm}
	\renewcommand{\algorithmicrequire}{\textbf{Input:}}
	\renewcommand{\algorithmicensure}{\textbf{Output:}}
	\caption{Process of Actor}
	\begin{algorithmic}[1]
		    \STATE Initialize environment ENV;
		    \STATE Initialize model $M$ with random parameters;
            \FOR{Episodes=1,2,3,...}
		        \STATE Initial state $s_{0}$ = ENV.reset();
		        \STATE Set $t = 0$;
		        \WHILE {not done}
    		        \FOR{Agent=1,2,3,4}
    		            \STATE $\tau^i_t=f(\tau^i_{t-1}, a^{-i})$;
        		        \STATE calculate the legal actions set $A$;
        		        \STATE choose action $a^i$ by $\epsilon$-greedy;
        		    \ENDFOR
                    \STATE $t = t+1$;
                \ENDWHILE
                \STATE Assign a value $r$ for every sample;
                \STATE For each trajectory $(\tau_t, a_t,Q(\tau_t,a_t), r_t)$, save it to replay buffer $B$;
                \STATE Update model $M$ with period $I$;
		    \ENDFOR
	\end{algorithmic}
	\label{actor}
\end{algorithm}

\begin{algorithm}
	\renewcommand{\algorithmicrequire}{\textbf{Input:}}
	\renewcommand{\algorithmicensure}{\textbf{Output:}}
	\caption{Process of Learner}
	\begin{algorithmic}[1]
        \STATE Initialize the network parameters and replay buffer $B$;
        \FOR{Iteration=1,2,3,...}
        \STATE Sample a batch of trajectory data $D=\{(\tau, a, Q(\tau,a), r)\}$ from $B$;
    	\STATE Calculate loss $L(\theta)$ as Eq.~\eqref{Target} and Eq.~\eqref{loss},
    	\STATE Update state-action value network parameters $\theta$ with $L(\theta)$;
        \STATE Send network parameters to Actor;
        \ENDFOR
	\end{algorithmic}
	\label{learner}
\end{algorithm}

\section{Experiment}
In this section, we compare the performance of DanZero with state-of-the-art rule-based methods on the GuanDan benchmark.
Also, in order to more intuitively demonstrate the strength of our AI, we test DanZero against human players. Our AI system is trained on a server with 4 Intel(R) Xeon(R) Gold 6252 CPU @ 2.10GHz and GeForce RTX 3070 GPU in Ubuntu 16.04 operating system.
The code is available at supplement.

\subsection{Experimental Setup}

In order to evaluate the performance of our AI program, we launch tournaments between our model and different baseline algorithms. To be specific, two agents of one team in GuanDan game utilize our model and another team adopts baseline algorithm. By initializing different games and executing the test for many times, we can achieve objective evaluation results.

We perform a mild hyper-parameter search on Q-learning and use the best setting for the shared hyper-parameters for all methods. An overview of hyper-parameters for each method is listed in the appendix. What's more, we adopt 80 actors for training.

\subsection{Performance against 8 Rule-based Bots and Each Other}
To show how our model performs in the training process, we save a checkpoint every 24 hours.
We evaluate these checkpoints by playing 1000 games with 8 rule-based bots. Considering that one GuanDan game contains several rounds, playing 1000 games is enough to reveal the performance of an AI program objectively.
To be mentioned, the baseline rule-based bots that we compare are the top 8 agents in the first Chinese Artificial Intelligence for GuanDan Competition so that their strengths can be guaranteed. 
Their implementations are available at this website \footnote{http://gameai.njupt.edu.cn/gameaicompetition/guandan\_machi\\ne\_code/index.html}.

\begin{figure}[t]
	\centering
	\includegraphics[width=0.96\columnwidth]{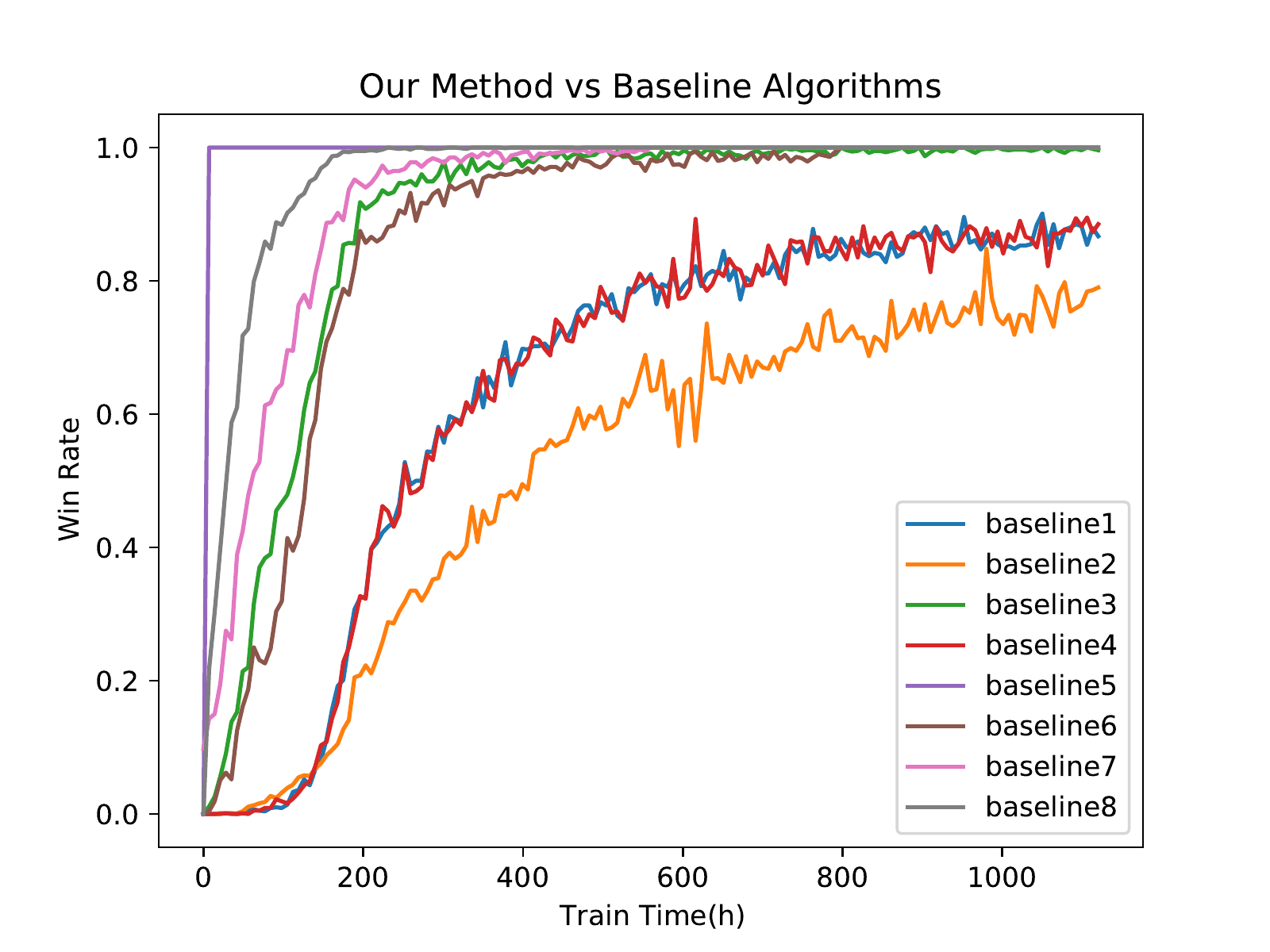}
	\caption{Winning rate for our algorithm against 8 rule-based methods. The horizontal axis represents training time of models and the vertical axis indicates the win rate of our model against rule-based bots. Evaluation against each baseline is executed with 1000 games every 24 hours.
	}
	\label{fig5}
\end{figure}

The average win rate of our model against different baselines is shown in Figure~\ref{fig5} and the indexes of baselines represent their ranks in the competition.
It can be observed that our AI system DanZero is able to achieve significantly better performance than those rule-based agents.
Specifically, DanZero has absolute superiority over the rule-based bots except for baseline 1, baseline 2 and baseline 4 after enough training.
What's more, considering that the nature of high variance of this game, achieving the win rate of 80\%  is an obvious preponderance so that our DanZero also performs much better than the other three baseline algorithms.

It's interesting that baseline 2 achieves the best result against our model instead of baseline 1 and baseline 3 seems to perform worse than baseline 4 and baseline 6. To figure out this phenomenon, we also conduct evaluations between all the AI programs and the results are shown in Table~\ref{tab1}. It can be observed that the performance of different baselines does not quite correspond to their ranks after enough evaluation, which can account for the above phenomenon partly. However, the overall performance of baseline 1, baseline 2 and baseline 4 is equivalent but baseline 2 obviously performs better than the other two against DanZero.
We assume that this is because there exists a restraint relationship between different policies.
In other words, a weak bot is also possible to beat a strong bot if its policy is just suitable to target the opponent's weakness.
This phenomenon is very common in AI programs using heuristic rules in fact.
However, our DanZero can still deal with these agents which have different styles and achieve obvious superiority, proving the effectiveness of our methods.

In addition, we also test the effectiveness of flag for ``wild cards" in the state feature. We conduct abrasive experiment that removes these dimensions in state features and Table~\ref{tab1} reports the results, which is represented by ``Our-". It can be observed that removing these information degrades the performance of our model, proving that adding this feature helps the model better grasp the use of ``wild cards".

\subsection{Human Evaluation}
Apart from comparisons with strong rule-based bots, we also evaluate The real performance of DanZero against human players.
After training for 30 days, DanZero plays over 100 games against 10 GuanDan players and achieves an average win rate of $60\%$.
Among games with human players, we find that DanZero makes good decisions in some cases.
In order to better illustrate the performance of our AI system, we list some case studies below in Figure~\ref{fig6}. 

In the first case of Figure~\ref{fig6}, it is player4's turn to play cards and the legal action set consists of Single, Pair and Straight. Player1 and player2 has only one card in hand and Player3 has two cards left. As the state feature contains the remaining cards so that Player4 knows that there are one 10, Q and two K left. In this case, playing Single 4, Pair 8 and Straight is all highly possible to result in failure so that the agent chooses to dismantle the Straight and play Single 10 to help the winning of partner.

In the second case, it's player2's turn to play cards and the legal action set is composed of Pair, Full House and so on.
Even if the teammate only has three cards in hand, player2 does not sacrifice himself to obtain the winning of his partner.
Meanwhile, player3 still has many cards in hand and the agent knows that there are quite a few 7 and 8 left, indicating that player3 is likely to have bombs in hand so player2 also does not play Red Joker.
At this time, player2 chooses to play Full House and keep Red Joker in hand. As there are also a few Single cards left, it's relatively easy to play the Red Joker out and win the game.

From the above discussed cases, it can be observed that DanZero learns when to cooperate and when to play ``selfishly'' for the winning, which is a high-level skill in this game, proving the strength of our model. 
\begin{figure}[t]
	\centering
	\includegraphics[width=0.91\columnwidth]{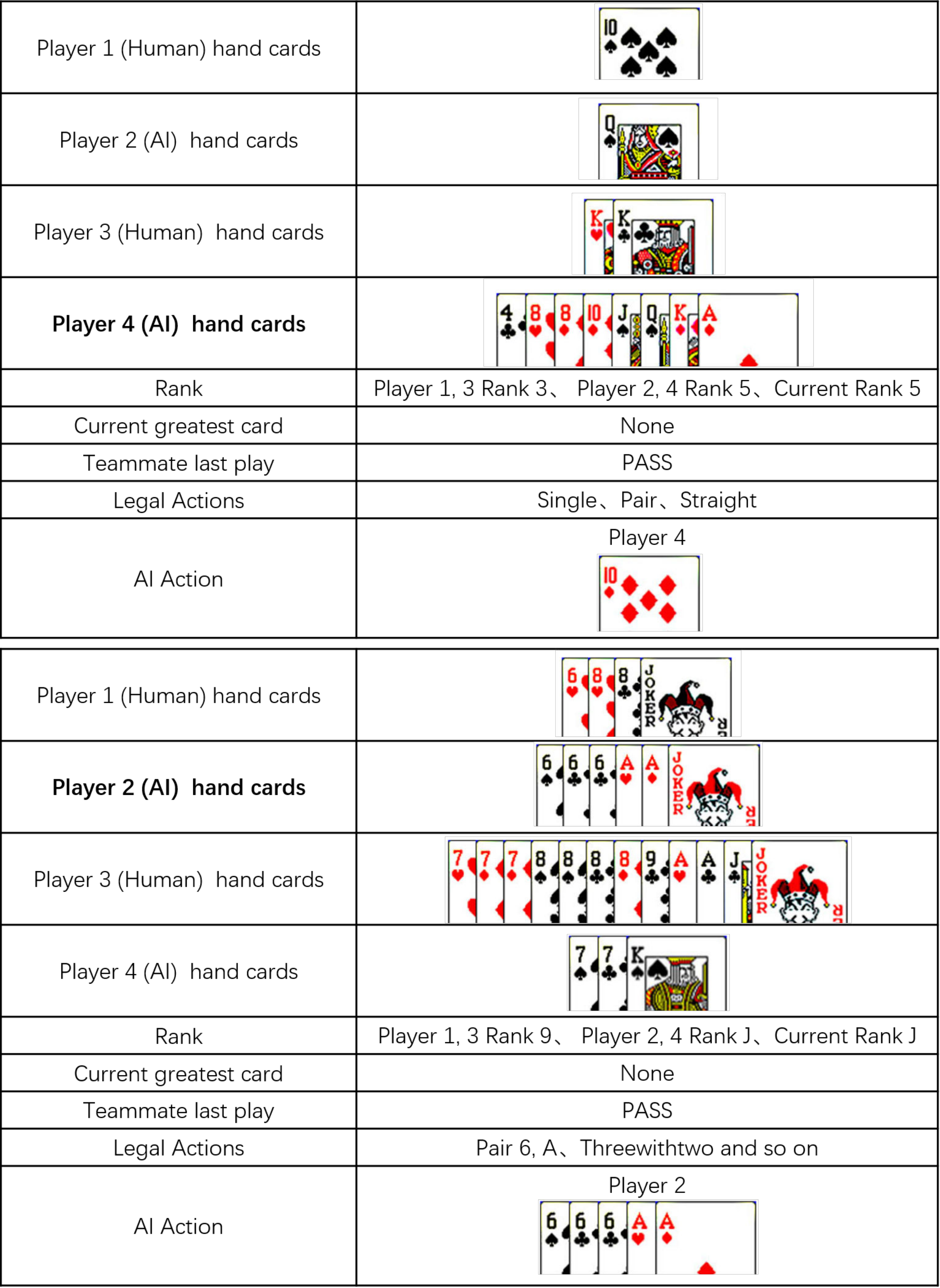}
	\caption{Case study to show skills of DanZero. The player 1 and player 3 are human players and player 2 and player 4 are controlled by our DanZero. The bold font indicates that it is the player's turn to play cards. The row of Rand indicates the levels of both teams and current round. The current greatest card means the last move in this trick and the cards that current player is going to play must cover it. The table also reports the last move of partner and legal action set. The row of AI Action is the actual action that the bot decides to take.}
	\label{fig6}
\end{figure}

\section{Conclusion}
In this paper, we propose an AI system for GuanDan, which is a very challenging imperfect-information game. In order to deal with the challenges such as large state and action space and unsure number of players, we adopt Deep Monte-Carlo Methods as the main algorithm, characterize the state information and utilize distributed self-play paradigm, leading to a strong RL bot, named DanZero. We compare our AI program with the state-of-the-art rule-based baselines and the outstanding performance reveals the effectiveness of our method. In addition, we execute human evaluation and DanZero manages to reach human level. We hope this work can be a benchmark for future research of GuanDan game.

For future work, we will try to enhance our AI system. Currently, our model takes a lot of time to train and we hope to find a method to accelerate the training process. What's more, the policy of DanZero in Tribute phase is based on heuristic rules and we will try reinforcement learning technique so that we can get a pure RL solution in this game.
\bibliographystyle{IEEEtran}
\bibliography{refer.bib}

\end{document}